\documentclass[11pt]{article}

\usepackage{epsfig}
\usepackage{amssymb}
\usepackage{natbib}
\usepackage{hyperref}
\usepackage{graphicx}
\usepackage{thumbpdf}
\usepackage{amsfonts,amstext,amsmath}
\usepackage{accents}
\usepackage{color}
\usepackage{rotating}
\usepackage[utf8]{inputenc}
\usepackage{booktabs}
\usepackage{tikz}
\usepackage{multirow}
\usepackage{float}
\usepackage[labelfont=bf, width=15cm, font=small]{caption}
\usepackage[onehalfspacing]{setspace}
\usepackage[left=3cm, right=3cm, top=2.5cm, bottom=3cm]{geometry}
\setcitestyle{authoryear,open={(},close={)}}
\setcitestyle{authoryear,open={(},close={)}}
\newcommand*{\doi}[1]{\href{http://dx.doi.org/#1}{doi: #1}}

\usepackage{dsfont}
\newcommand{\I}{\mathds{1}}
\newcommand{\dd}{\mathrm{d}}
\newcommand{\ie}{\textit{i.e.},~}
\newcommand{\eg}{\textit{e.g.},~}

\newcommand{\cf}{\textit{cf.}~}

\newcommand{\given}{\mid}
\DeclareMathOperator{\NLL}{NLL}

\newcommand{\dlsa}{DCTM}
\newcommand\blfootnote[1]{%
  \begingroup
  \renewcommand\thefootnote{}\footnote{#1}%
  \addtocounter{footnote}{-1}%
  \endgroup
}

\begin{document}

\title{\bf Deep conditional transformation models for\\survival analysis}

\author{Gabriele Campanella\textsuperscript{1,2,*}, 
        Lucas Kook\textsuperscript{3,4,*},
        Ida H\"aggstr\"om\textsuperscript{5,6}, \\
        Torsten Hothorn\textsuperscript{3},
        Thomas J. Fuchs\textsuperscript{1,2}
        \blfootnote{Authors contributed equally.\newline Corresponding author: 
        \texttt{thomas.fuchs.ai@mssm.edu}\newline Preprint. Version October 20, 2022.
        Licensed under CC-BY.}
        }

\date{\footnotesize%
\textsuperscript{1}Department of AI and Human Health, Icahn School of Medicine at Mount Sinai, New York, 10029, USA \\
\textsuperscript{2}Hasso Plattner Institute for Digital Health at Mount Sinai, New York, 10029, USA \\
\textsuperscript{3}Epidemiology, Biostatistics \& Prevention Institute, University of Zurich, CH-8001, Switzerland \\
\textsuperscript{4}Institute for Data Analysis and Process Design, Zurich University of Applied Sciences, CH-8400, Switzerland \\
\textsuperscript{5}Chalmers University of Technology, Department of Electrical Engineering, 41296, Sweden \\
\textsuperscript{6}Memorial Sloan Kettering Cancer Center, Department of Radiology, New York, 10065, USA \\
}

\maketitle

\begin{abstract}%
An every increasing number of clinical trials features a time-to-event outcome and records non-tabular patient data, such as magnetic resonance imaging or text data in the form of electronic health records. Recently, several neural-network based solutions have been proposed, some of which are binary classifiers. Parametric, distribution-free approaches which make full use of survival time and censoring status have not received much attention. We present deep conditional transformation models (\dlsa{s}) for survival outcomes as a unifying approach to parametric and semiparametric survival analysis. \dlsa{s} allow the specification of non-linear and non-proportional hazards for both tabular and non-tabular data and extend to all types of censoring and truncation. On real and semi-synthetic data, we show that \dlsa{s} compete with state-of-the-art DL approaches to survival analysis.
\end{abstract}

\section{Introduction} \label{sec:introduction}

Arguably one of the most important aspects of health and medical research is 
being able to understand and predict patient outcome in order to improve 
patient management and ultimately extend their life span or time in remission
\citep{hosny_deep_2018}.
Survival analysis is used for these purposes to study time-to-event information 
relating to for example death, response to treatment, adverse treatment effects,
disease relapse, and the development of new disease \citep{collett2015modelling}.
Traditional approaches, such as Cox Proportional Hazards (\cf Section~\ref{sec:relatedwork}),
relied on tabular features and are not amenable to analyze high-dimensional non-tabular data 
such as medical images. With recent advances in computer vision and deep leaning, there has been increasingly
more interest in performing survival analysis directly from high-dimensional data in order to
automatically learn patterns that stratify patients based on their outcome without the need 
for feature engineering.

In this paper we present DCTM, a framework for parametric and semiparametric survival analysis 
rooted in statistical modeling. \dlsa{s} allow the specification of non-linear and non-proportional 
hazards for both tabular and non-tabular (image or text) data. We will describe in 
detail the formalization of our proposed models and apply them to a real dataset of medical images.
Additionally, we describe how DCTMs can be used as generative models for the generation of 
semi-synthetic data. To the best of our knowledge, this paper is the first to cover deep survival 
regression from the DCTM point of view.

\section{Deep transformation models for survival outcomes} \label{sec:method}

\begin{figure}[!ht]
    \centering
    \includegraphics[width=\textwidth]{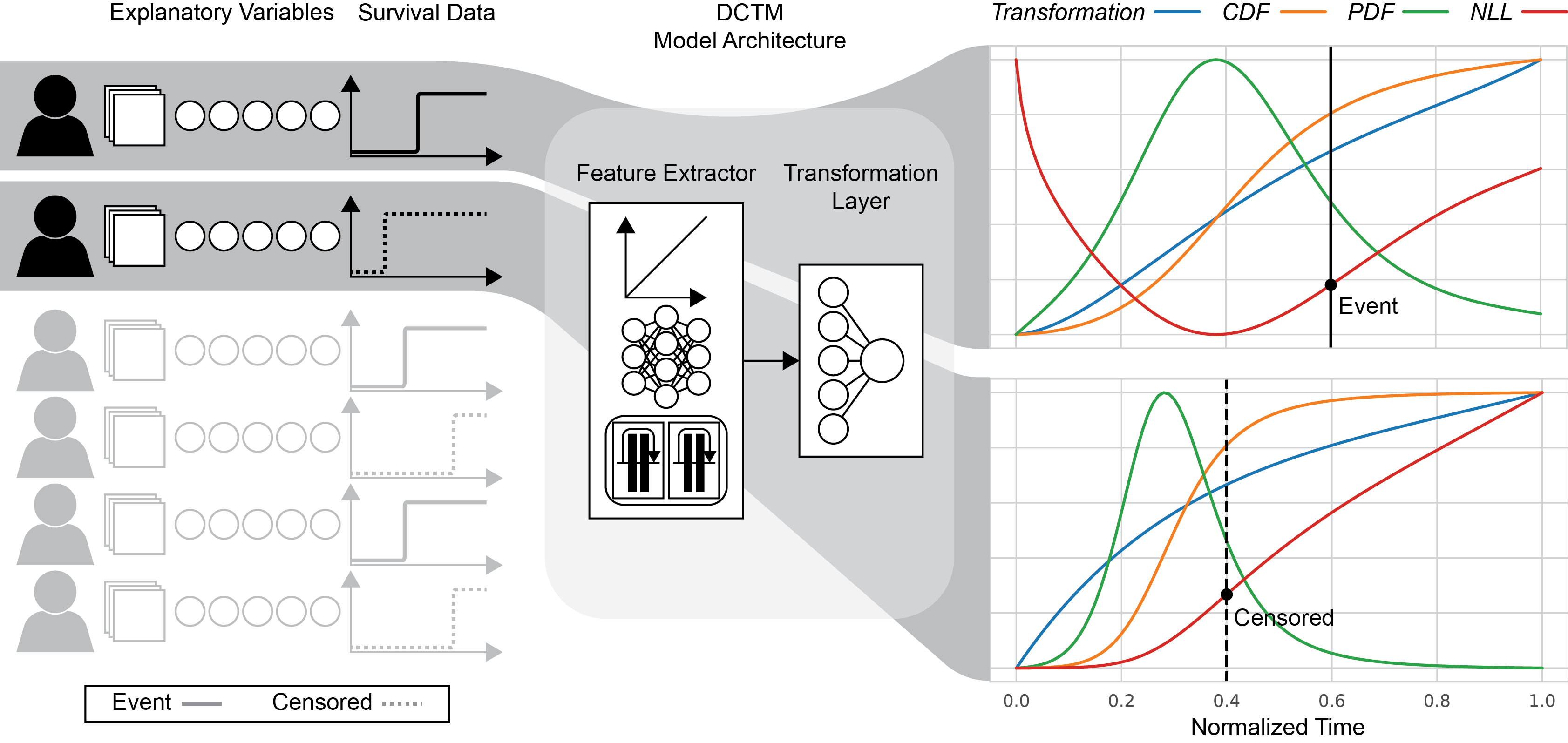}
    \caption{Proposed DCTM model architecture. The inputs to the model consist of 
    tabular or non-tabular explanatory variables associated with survival data 
    in terms of exact or right-censored times to event. The explanatory variables 
    can be mapped to a latent feature space via a feature extractor. In the case 
    of non-tabular data such as images, the feature extractor will consist of a 
    non linear mapping such as a convolutional neural network. The extracted 
    features are then fed to the transformation layer. The network is optimized by 
    minimizing the negative log-likelihood (NLL). In the figure we show the 
    transformation function along with the cumulative density function (CDF),
    probability density function (PDF) and NLL for an exact event and a 
    right-censored example.}
    \label{fig:method}
\end{figure}

In the following, we introduce deep conditional transformation models (DCTMs) for survival 
outcomes. We briefly recap survival analysis and conditional transformation models.
Then we describe how to setup, fit, evaluate and sample from our proposed models.
Fig.~\ref{fig:method} is an overview of the proposed class of models.

\paragraph{Survival analysis}
Survival analysis characterizes the distribution of the positive real-valued event 
time $T$ conditional on covariates $X$, usually on the scale of the survivor function,
$S_{T \given X = x}(t) = 1 - F_{T \given X = x}(t)$, where $F_{T \given X = x}$ denotes the 
cumulative distribution function (CDF) of $T \given X = x$. One of the most popular choices
is the Cox proportional hazards (Cox PH) model \citep{cox1972regression}
\begin{align} \label{eq:coxph}
    F_{T \given X = x}(t) = 1 - \exp(-\Lambda(t \given x)) = 
    1 - \exp(-\Lambda_0(t)\exp(x^\top\beta)),
\end{align}
where $\Lambda(t \given x)$ denotes the positive, monotone increasing cumulative hazard 
function. The hazard function is assumed to be decomposable into a baseline hazard $\Lambda_0$,
independent of the covariates, and a hazard ratio $\exp(x^\top\beta)$. The hazard ratio models 
the influence of the covariates on the survivor function. The hazard function 
$\lambda(t \given x) = \frac{\dd}{\dd t} \Lambda(t \given x)$ measures the instantaneous risk
of an event after time $t$ conditional on the covariates $x$ and having survived until $t$
\citep{collett2015modelling}.

The Cox PH model is commonly estimated using the maximum partial likelihood obtained from
profiling out $\Lambda_0$ \citep{cox1975partial}. The connection to transformation models 
for distributional regression is drawn next.

\paragraph{Conditional transformation models}
CTMs are parametric distributional regression models of the form 
\citep{hothorn2014conditional}
\begin{align} \label{eq:ctm}
    F_{T\given X = x}(t) = F_Z(h(t \given x)),
\end{align}
where the conditional distribution of the response $T \given X = x$ is decomposed into 
an \emph{a priori} chosen and parameter-free target distribution $F_Z$ and a conditional
transformation function $h$, which depends on the input data $x$. In order for $F_T$
to be a valid CDF, $h$ needs to be monotonically increasing in $t$ \citep{hothorn2018most}.

CTMs can be estimated via maximum likelihood and allow various kinds of responses
and uninformative censoring. The model in \eqref{eq:ctm} suggests a close connection
to the Cox PH model in \eqref{eq:coxph}, namely for $F_Z(z) = 1 - \exp(-\exp(z))$ 
(the minimum extreme value distribution) and $h(t \given x) = \log\Lambda(t \given x)$, 
the two models coincide.

\paragraph{DCTMs for survival analysis}
In DCTMs, the transformation function is parameterized via (deep) neural networks.
For instance, let $\phi: \mathcal{X} \to \mathbb{R}^d$ denote a feature extractor,
which maps the input $x$ to a feature vector of dimension $d$. We can then choose 
different parameterizations for the transformation function, depending on the desired 
complexity of the model. For example,
\begin{align}
    h(t \given x; \phi) = \alpha + \beta\log(t) + \phi(x)^\top w, \quad \beta > 0,
\end{align}
together with $F_Z(z) = 1 - \exp(-\exp(z))$ is a Weibull proportional hazards model
with non-linear log hazard-ratios depending on the input $x$ (for instance, medical images).
However, also non-proportional (time-varying) hazards can be realized in DCTMs, via
\begin{align}
    h(t \given x; \phi) = \alpha + g(\phi(x)^\top w) \log(t),
\end{align}
where $g: \mathbb{R} \to \mathbb{R}_+$, \eg the soft-plus function, ensures a valid CDF.
In Section~\ref{sec:param}, we describe the parameterization of the DCTMs used in this
paper. An overview of the proposed method is given in Figure~\ref{fig:method}.

A parametric version of the Cox PH model can be estimated in the DCTM framework.
Here, the log cumulative baseline hazard function is estimated as a smooth basis expansion
$a(t)^\top\vartheta$, instead of using the non-parametric estimate. A common choice
are Bernstein polynomials, which are easily constrained to be monotonically increasing
(see Section~\ref{sec:experiments}).

\paragraph{Fitting DCTMs}
DCTMs are fitted by optimizing the empirical negative log-likelihood (NLL) via stochastic 
gradient descent (SGD). The likelihood contribution of a single observation $(t, x)$ can 
be expressed in terms of the general transformation model
\begin{align}
    \mathcal{L}(h; t, x) = \begin{cases} 
        f_Z(h(t \given x))h'(t \given x) & t \mbox{ exact event}, \\
        1 - F_Z(h(t \given x)) & t \mbox{ right censored}.
    \end{cases}
\end{align}
However, also interval- and left-censored observations can be handled with the proposed
method \citep[see \eg][]{hothorn2018most}. For a general choice of the transformation
function $h(t \given x)$ and error distribution $F = \sigma$, we have the following likelihood 
function
\begin{align}
    \mathcal{L}(h; t, x) = \begin{cases} 
        \sigma(h(t \given x)) (1 - \sigma(h(t \given x)) h'(t \given x) & t \mbox{ exact event}, \\
        1 - \sigma(h(t \given x)) & t \mbox{ right censored}.
    \end{cases}
\end{align}
Lastly, the NLL
\begin{align} \label{eq:nll}
    \NLL = - \sum_{i=1}^{n_t} \log \mathcal{L}(h; t_i, x_i),
\end{align}
is minimized via SGD, where $n_t$ denotes the training sample size.

\paragraph{Evaluating DCTMs}
After fitting a DCTM, the conditional survivor function of a test observation can be computed from
the estimated parameters via $\hat{F}_{T \given X = x}(t) = F_Z(\hat h(t \given x))$. Now, all 
evaluation metrics can be computed from (some form of) the predicted conditional distribution.

Evaluating models that predict conditional distribution of survival times are not straight-forward
to evaluate, due to censoring and their probabilistic nature \citep{collett2015modelling}. For instance, 
mean squared error or median absolute deviation are insensible for a right-censored observation 
$((t, +\infty], x)$ and prediction $\hat t$, for instance the conditional median survival time.

The c-index \citep{harrell1982cindex}
\begin{align}
    c = \frac{\sum_{i,j}\I_{T_j<T_i}\cdot 
    \I_{\eta_j>\eta_i} \cdot \delta_j}{\sum_{i,j}\I_{T_j<T_i}\cdot \delta_j},
\end{align}
overcomes at least the issue of censoring. However, the c-index does not encourage faithful probabilistic
predictions, because it is not a proper scoring rule \citep{blanche2018cindex}. 

Proper scoring rules \citep{gneiting2005weather,gneiting2007strictly} are explicitly designed to
evaluate probabilistic predictions and can inherently deal with censoring. A score is proper, if
its average w.r.t. the data-generating distribution is minimized when predicting exactly this 
distribution, \ie if $T \sim \mathbb{P}_{T \given X = x}$,
\begin{align} \label{eq:proper}
    \mathbb{E}_{\mathbb{P}_{T \given X = x}}[S(p; T)] \leq \mathbb{E}_{\mathbb{P}_{T \given X = x}}[S(q; T)].
\end{align}
Here, $\mathbb{P}_{T \given X = x}$ denotes the data-generating probability distribution with density $p$ and 
$q$ is another predicted density. A score is called \emph{strictly proper}, when the above inequality holds if 
and only if $p = q$.

The c-index does not fulfill \eqref{eq:proper}, whereas the negative log-likelihood 
\begin{align} \label{eq:logscore}
   S_{\mathrm{NLL}}(\hat F, t) = - \delta \log \hat f(t) - (1 - \delta) \log \hat S(t)
\end{align}
does \citep{good1952rational}. Here, $\delta \in \{0, 1\}$ denotes an exact ($\delta = 1$) or a censored 
($\delta = 0$) survival time. In fact, the NLL is the only strictly proper local scoring rule
\citep[up to affine transformations,][]{brocker2007scoring}.
However, since DCTMs are fitted by minimizing the empirical NLL, a comparison based solely on this 
score could be deemed unfair, if the competing method optimizes a different score, say the c-index.
For this reason, we also compute the continuous ranked probability score \citep[see \eg][]{avati2020crps}
\begin{align} \label{eq:crps}
   S_{\mathrm{CRPS}}(\hat F, t, \delta) = 
       \int_0^t \hat F^2(u) \; \dd u + \delta \int_t^\infty (1 - \hat F(u))^2 \; \dd u.
\end{align}

\begin{figure}[!ht]
    \centering
    \includegraphics[width=0.95\textwidth]{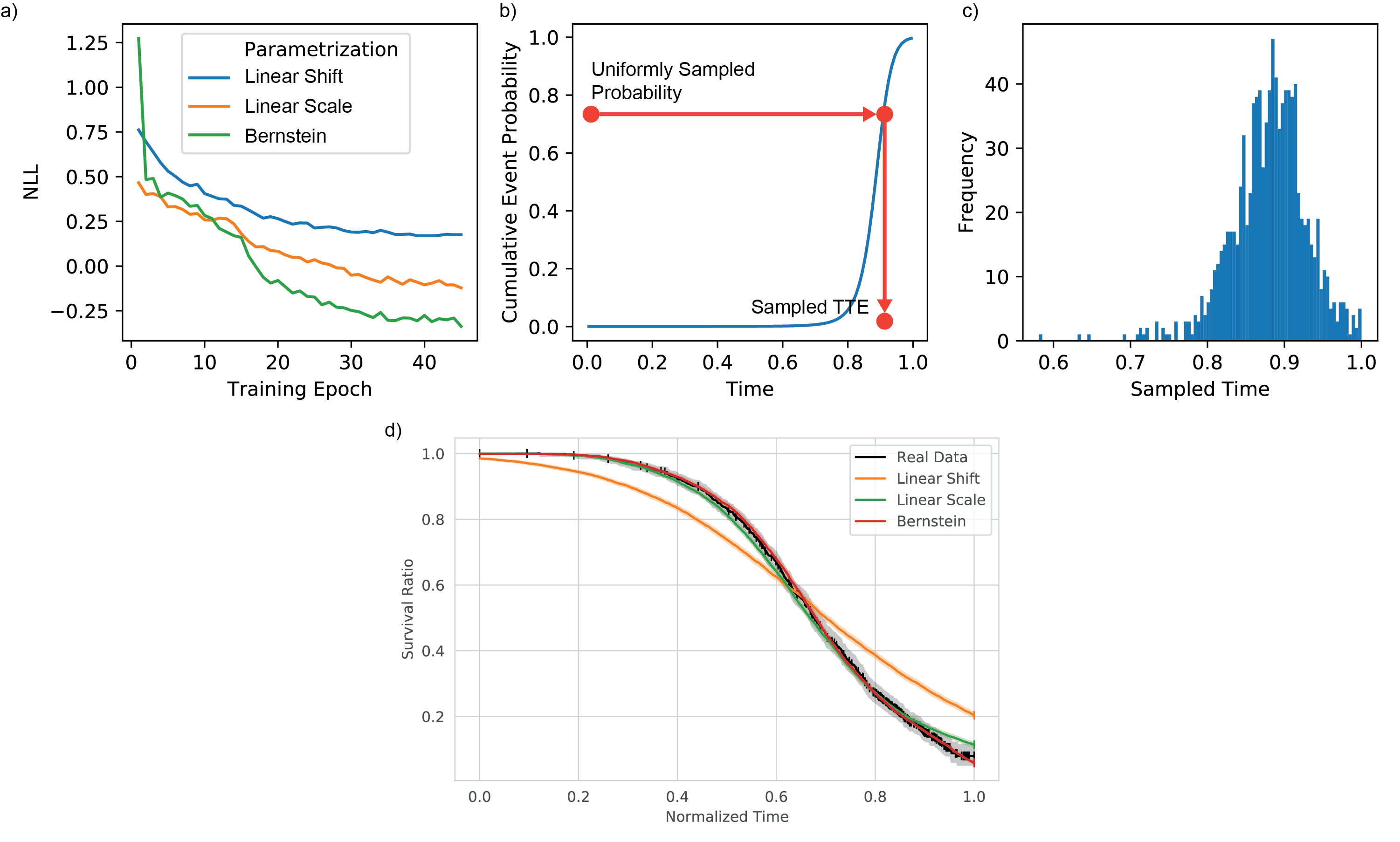}
    \caption{Generation of semi-synthetic data via inversion sampling for DCTMs. a) Fitting of
    real data with three DCTM parametrizations of increasing complexity. The more flexible models 
    reach a lower NLL. b) Example of a patient's fitted CDF and one sampled synthetic time. 
    c) Distribution of 1,000 sampled times for the previous CDF. d) Generated semi-synthetic 
    datasets based on the three DCTM parametrizations compared to the real data.}
    \label{fig:sampling}
\end{figure}

\paragraph{DCTMs as generative models} 
DCTMs model the entire conditional distribution of $T \given X = x$. Consequently, one can sample from
the estimated distribution $\hat F_{T \given X = x}(t)$, \eg via inversion sampling (see Figure~\ref{fig:sampling}).
In more detail, given a real sample and a fitted DCTM, the sample's CDF can be generated. Next, random uniform 
probability values can be drawn and mapped back to time via the inverse CDF (Figure~\ref{fig:sampling}b).
The usefulness of sampling from DCTMs is illustrated in our semi-synthetic experiments
in Section~\ref{sec:results}.

\section{Related work} \label{sec:relatedwork}

We summarize the concurrent literature on deep learning for survival analysis and
transformation models.

\paragraph{Deep learning for survival analysis}
We distinguish Bayesian and frequentist approaches and the loss function involved in fitting 
the models.

Several DL approaches to survival analysis are inspired by the Cox PH model.
One of the earliest approaches using (shallow) neural networks for survival analysis is
\citet{Liestbl1994}. The authors use the partial likelihood as a loss function and
even describe extensions to piece-wise constant, or non-proportional hazards and non-linear 
effects. 
A direct extension to deep Cox PH models is ``DeepSurv'' \citep{katzman2018deepsurv}, which
optimizes the $\ell_2$-regularized log partial-likelihood.
Also, \citet{Nagpal2021} proposes a mixture of Cox models together with an expectation 
maximization (EM) algorithm to optimize the partial likelihood and the evidence lower bound 
(ELBO) of a variational auto-encoder for the features.

In contrast to the semi-parametric approaches derived from the Cox PH model, 
a fully parametric framework for survival analysis was given in \citet{nagpal2020deep}.
The authors model the survivor function as a mixture distribution and fit the model 
using an ELBO for exact and censored responses. 
\citet{Ranganath16} present a generative (Bayesian) Weibull model for survival times.
Some approaches based on pseudo-values exist \citep{zhao2020dnnsurv,rahman2020pseudo}. 
\citet{Fornili2014} employ piece-wise exponential hazard functions in a penalized
likelihood loss. 

``DeepHazard'' proposed by \citet{Rava2020}, uses a squared error loss based on the
underlying counting process, which takes into account censoring. The model allows for
time-varying covariates and non-linear effects, as well as non-proportional hazards.

A different branch of literature treats survival regression as a classification problem.
\citet{fotso2018deep} introduce ``neural multi-task logistic regression'' (N-MTLR),
which splits the positive real line into $K$ sub-intervals and fits a softmax last-layer
model to the event indicators. The authors evaluate their model using proper scoring rules,
namely the weighted and integrated Brier score (which is equivalent to the CRPS).
\citet{lee2019dynamic} propose ``dynamic DeepHit'', which is based on ``DeepHit''
\citep{lee2018deep} and again uses a softmax last-layer activation. (Dynamic) DeepHit is able 
to deal with competing risks and non-linear effects.

DL for survival analysis has been shown to be useful in medical applications. For instance,
\citet{Lao2017} use DL together with radiomics features for prediction of glioblastoma 
multiforme survival. A comparison between a deep learning and the classical Cox PH 
can be found in \citep{Matsuo2019}. DeepSurv has been applied and compared against the Cox PH
model and survival random forests for oral cancer \citep{kim2019deepsurvapplication}
and non-metastatic clear cell renal cell carcinoma \citep{byun2021deepsurvapplication}
survival prediction.
    
\paragraph{Deep conditional transformation models}
Our proposed class of models belongs to the family of DCTMs. DCTMs extend the flexible class 
of CTMs \citet{hothorn2014conditional,hothorn2018most} with deep neural networks and SGD to
handle non-tabular data, such as images. TMs are distributional regression models, which
capture the entire conditional distribution instead of a single or few moments thereof
\citep{kneib2021rage}. More detail on TMs in given in Section~\ref{sec:method}.

DCTMs were first introduced by \citet{sick2020deep}, where the authors used a series
of nested functions to model the transformation function and the standard normal CDF as the
target distribution. \citet{baumann2020deep} extended the work on DCTMs for continuous
distributions with exact observations. \citet{kook2020ordinal} treated ordinal regression
from the DCTM point of view. The authors view the ordinal response as an interval censored
version of a latent logistic distribution. By imposing structural assumptions on the
transformation function, such as additivity, DCTMs retain interpretability of certain parts
of the models, \eg parameters for a single data modality. DCTMs were also developed for 
time series, \ie serially correlated data \citep{rugamer2021timeseries}.
DCTMs are all fitted by minimizing the empirical NLL via some form of SGD.

\section{Experimental setup} \label{sec:experiments}

In the following, we describe the details of the experimental setup
including parameterization of the DCTMs and the use of ensemble estimates for
improving prediction performance.

\subsection{Model parametrizations} \label{sec:param}

We employ different parameterizations of the transformation function, namely a
linear shift, a linear scale, and general shift and scale models with a non-linear 
function in $t$ using Bernstein polynomials. 
For all experiments, we choose the standard logistic CDF $\sigma(z) = (1 + \exp(-z))^{-1}$
or the standard minimum extreme value CDF (also called Gompertz) $F_{\mathrm{MEV}}(z) = 1 - \exp(-\exp(z))$
as the target distribution. Consequently, the transformation function can be interpreted
on the log-odds or log-hazard scale, respectively.

For the linear shift parameterization, the transformation function is parameterized as 
\begin{equation}
    \begin{split}
        h(t \given x; \phi) = a + b\log t + \phi(x)^\top w, \mbox{ where} \\
        \phi: \mathcal{X} \to \mathbb{R}^P, a \in \mathbb{R}, b \in \mathbb{R}, w \in \mathbb{R}^P,
    \end{split}
\end{equation}
where $h$ denotes the transformation function, with a linear basis in $t$ parameterized via
$a$ and $b$, and a feature extractor $\phi$ representing a neural network with linear last-layer
activation and weights $w$.
Here, the extracted features enter linearly on the scale of the transformation function.
In turn, the conditional distribution function changes only in terms of location, but not
in terms of scale or any higher moments. That is, $F_{T \given X = x}$ is restricted to the 
same family of distributions as the target distribution $F$. Hence, this model is not 
distribution-free. For $F_Z = F_{\rm MEV}$, the linear shift model is equivalent to a Weibull 
model. Similarly, $F_Z = \operatorname{expit}$, yields a log-logistic model.

For the linear scale model, we have
\begin{equation}
    \begin{split}
        h(t \given x; \phi) = a + \operatorname{softplus}(\phi(x)^\top w) \cdot \log t, \mbox{ where} \\
        \phi: \mathcal{X} \to \mathbb{R}^P, a \in \mathbb{R}, w \in \mathbb{R}^P,
    \end{split}
\end{equation}
in which the neural network models the scale of $F_{T \given X = x}$, which allows for linear 
non-proportional hazards. Still, the model is not distribution-free, the resulting distributions 
are again Weibull or log-logistic.

The general shift transformation model parameterizes the transformation function using Bernstein
polynomials, 
\begin{equation}
    \begin{split}
        h(t \given x; \phi) = b(\log t)^\top\vartheta + \phi(x)^\top w, \mbox{ where}, \\
        \phi: \mathcal{X} \to \mathbb{R}^{P}, b: [0, 1] \to \mathbb{R}^{K + 1},
    \end{split}
\end{equation}
which requires $\vartheta$ to be increasing, \ie $\vartheta_{k+1} > \vartheta_k$, $k = 1, 
\dots, K$, for all $x$ to ensure monotonicity of $h(t \given x; \phi)$ in $t$ \citep{hothorn2014conditional}.
Monotonicity can be ensured by transforming 
\begin{align*}
    \vartheta = g(\gamma) = \left(\gamma_1, \gamma_1 + \operatorname{softplus}(\gamma_2),
    \dots, \gamma_1 + \sum_{k = 2}^{K+1} \operatorname{softplus}(\gamma_k) \right).
\end{align*}
The general shift transformation model allows a flexible baseline hazard function.
The transformation function influences all higher moments of $F_{T \given X = x}$ and the 
model is distribution-free. Thus, the general shift model can be viewed as a parametric 
version of the Cox proportional hazards model.

The shift-scale transformation model,
\begin{equation}
    \begin{split}
        h(t \given x; \phi) = \operatorname{softplus}(\phi(x)^\top \beta) \cdot 
        b(\log t)^\top\vartheta + \phi(x)^\top w, \mbox{ where} \\
        \phi: \mathcal{X} \to \mathbb{R}^{P}, b: [0, 1] \to \mathbb{R}^{K+1},
    \end{split}
\end{equation}
allows for non-proportional hazards, explicitly modelling the scale of $F_{T \given X = x}$.
The model is distribution-free \citep{siegfried2022distribution}.

Allowing the parameters of the Bernstein polynomials to fully flexibly depend on $x$, we
arrive at the most flexible transformation model
\begin{equation}
    \begin{split}
        h(t \given x; \vartheta) = b(\log t)^\top\vartheta(x), \mbox{ where} \\
        \phi: \mathcal{X} \to \mathbb{R}^{P}, b: [0, 1] \to \mathbb{R}^{K+1}
    \end{split}
\end{equation}

It is important to note that for all of the above $u = \log t$ is scaled to 
$\tilde u = \frac{u - a}{b - a}$ where $a, b$ are chosen appropriately, \eg min and max.

\subsection{Ensemble predictions} \label{sec:ensembling}

The linear shift and linear scale parameterizations induce strong
distributional assumptions on $T \given X = x$. Nevertheless, these distributional
assumptions can help prevent overfitting and lead to more stable predictions.
Using the Bernstein polynomial basis loosens assumptions, but is prone to
overfitting. However, deep ensembles have been shown to improve prediction
performance in both cases \citep{lakshminarayanan2016deepensembles}.

Let $\hat F_1, \dots, \hat F_M$ be $M$ estimates of the conditional distribution
of $T \given X = x$, \eg obtained by optimizing $M$ instances of the same DCTM via SGD
and early stopping on different bootstrapped data samples. The ensemble estimate is 
the point-wise average of those $M$ estimates, and denoted by $\bar F_M$,
\begin{align}
    \bar F_M = M^{-1} \sum_{m = 1}^M \hat F_m.
\end{align}
The ensemble distribution is then used for evaluation. When having access to $B > M$
estimates and selecting $M$ based on their validation loss, we refer to $\bar F_M$
as the top-$M$ ensemble estimate.

\subsection{Data} \label{sec:data}

\subsubsection{Real data}

The real data used in this study comprised 959 stage 4 lung cancer patients from Memorial 
Sloan Kettering Cancer Center (MSKCC) diagnosed from Oct 2010 through Mar 2018. The data 
used included pre-treatment PET/CT images of the subjects, overall survival time and 
censoring information. 768 of the patients (80\%) experienced a terminal event, while 191 
(20\%), were right-censored. The patients comprise a real-world, clinical grade cohort 
collected over many years on a wide selection of scanners, with cases reviewed by many 
different clinicians, and patients prescribed varying treatment regimens. The 
retrospective data use was approved by the local ethics committee at MSKCC, and 
informed consent was waived since the study was deemed minimum risk.

\subsubsection{Semi-synthetic data}
The DCTM architecture was used to model the conditional time to event distribution of the 
real data described above. Three different DCTM parametrizations were used of increasing 
complexity: linear shift, linear scale, and fully flexible Bernstein. The entire real 
data dataset was used to train each DCTM model for a predetermined number of epochs.
The fitted models were then used to generate new semi-synthetic time-to-event data. If a 
time was predicted beyond the real data's range, it was right-censored to the maximum 
value. For each patient we sampled 10 new data points, resulting in datasets 10 times 
larger than the original (see Figure~\ref{fig:sampling}d).

\subsection{Computational details} \label{sec:compdetails}

All experiments were performed on the high-performance computing (HPC) cluster at MSKCC. 
Training scripts were written in python using PyTorch \citep{paszke_automatic_2017} for 
model definition and optimization. Model optimization was performed with SGD and training 
learning rates were chosen via a hyperparameter grid search on a single data 
training/validation partition. For the DCTM models, different learning rates were used for 
the CNN feature extractor and the DCTM survival layer. The full list of learning rates used 
for all experiments is presented in \ref{tab:lr}.

\begin{table}[!ht]
\centering
 \caption{\label{tab:lr}List of learning rates used to train models for all experiments in this study.}
 \begin{tabular}{llrr} 
 \toprule
 \bf Parametrization & \bf CDF & \bf lr CNN & \bf lr DCTM \\
 \midrule
 Deepsurv & & 0.0005 & \\
 
 Baseline & Sigmoid & & 0.1\\
 Linear shift & Sigmoid & 0.001 & 0.01 \\
 Linear scale & Sigmoid & 0.001 & 0.1 \\
 Bernstein shift & Sigmoid & 0.001 & 0.1 \\
 Bernstein shift/scale & Sigmoid & 0.001 & 0.01 \\
 Bernstein & Sigmoid & 0.001 & 0.1 \\
 
 Baseline & Gompertz & & 0.01\\
 Linear shift & Gompertz & 0.001 & 0.01 \\
 Linear scale & Gompertz & 0.001 & 0.01 \\
 Bernstein shift & Gompertz & 0.001 & 0.01 \\
 Bernstein shift/scale & Gompertz & 0.001 & 0.01 \\
 Bernstein & Gompertz & 0.001 & 0.1 \\
 \bottomrule
 \end{tabular}
\end{table}

\section{Results} \label{sec:results}

\subsection{Real Data}

\begin{figure}[!ht]
    \centering
    \includegraphics[width=1\textwidth]{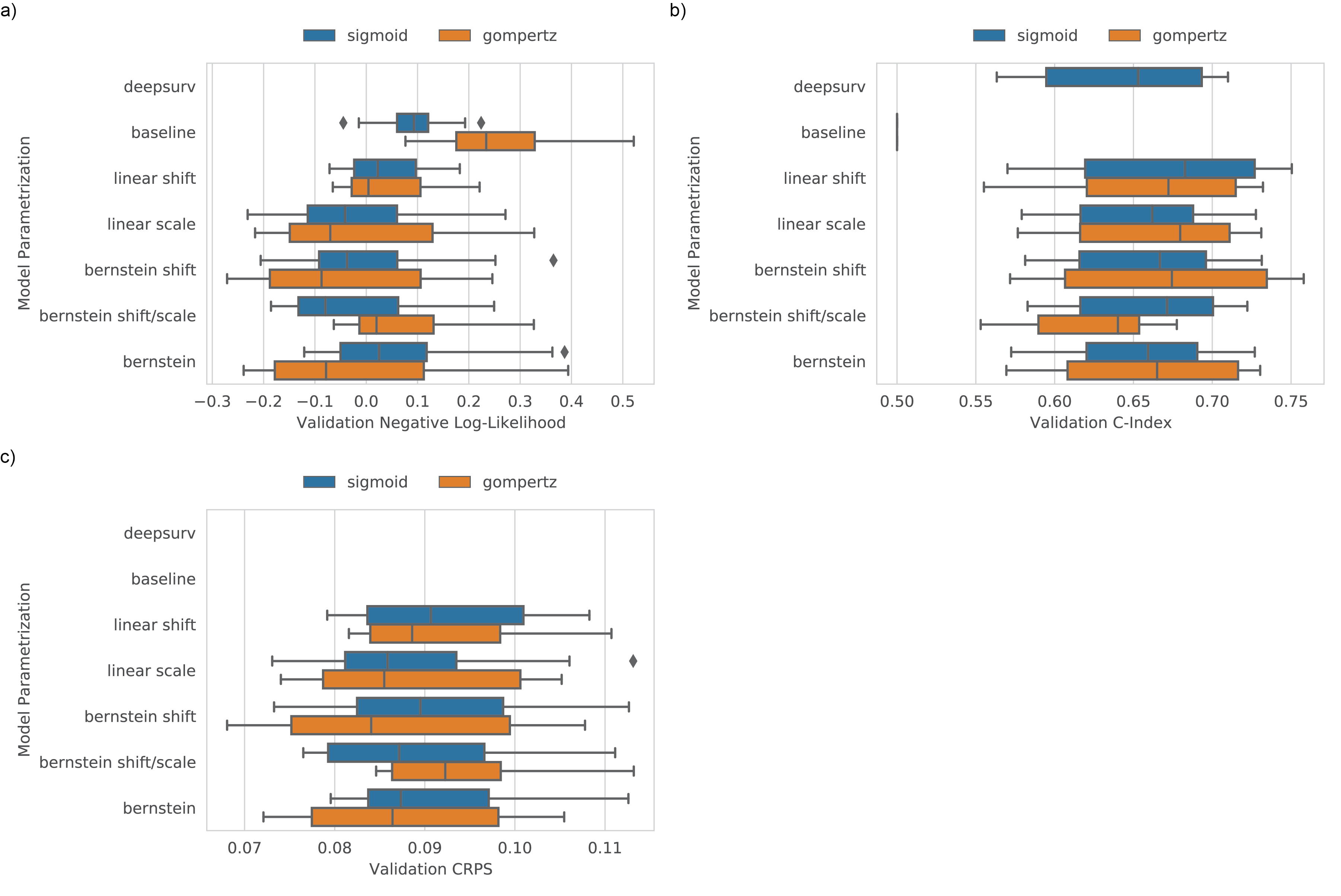}
    \caption{Experimental results for the real data. The performance of different parametrizations 
    of DCTM are shown compared to DeepSurv where possible. For DCTM, the sigmoid and Gompertz 
    functions were tested. a) Distribution of validation NLL. b) Distribution of validation 
    c-index. c) Distribution of validation CRPS.}
    \label{fig:results_real}
\end{figure}

Experiments were carried out to test the performance of DCTM models of increasing complexity. 
As a baseline a Bernstein non conditional model, where the same output is associated to every 
sample, was used. DCTM results were also compared with DeepSurv. Since DeepSurv does not 
compute the full likelihood, only the c-index can be compared in this case.
Figure~\ref{fig:results_real}a reports the distributions of validation NLL for each experimental 
condition in order of complexity, starting with the baseline at the top and finishing with the 
fully flexible Bernstein model at the bottom. It can be observed that, for both CDF functions, 
the NLL decreases with model complexity, as expected. In addition, it is apparent that the 
Gompertz experiments resulted in wider NLL 
distributions, highlighting the underlying training instability we observed for this subset of 
experiments.
Figure~\ref{fig:results_real}b reports the c-index results. The baseline results are by definition 
0.5, since all patients are assigned the same output time-to-event. It is interesting to note that 
while the NLL decreased with model complexity, a slight trend of decreasing c-index can be observed.
Compared to DeepSurv, the DCTM models perform on par or better, except in one condition.
Finally, Figure~\ref{fig:results_real}c reports the results in terms of CRPS. In concordance with 
the NLL results, a slight trend of decreasing CRPS with model complexity can be noted.

\subsection{Simulated Data}

\begin{figure}[!ht]
    \centering
    \includegraphics[width=1\textwidth]{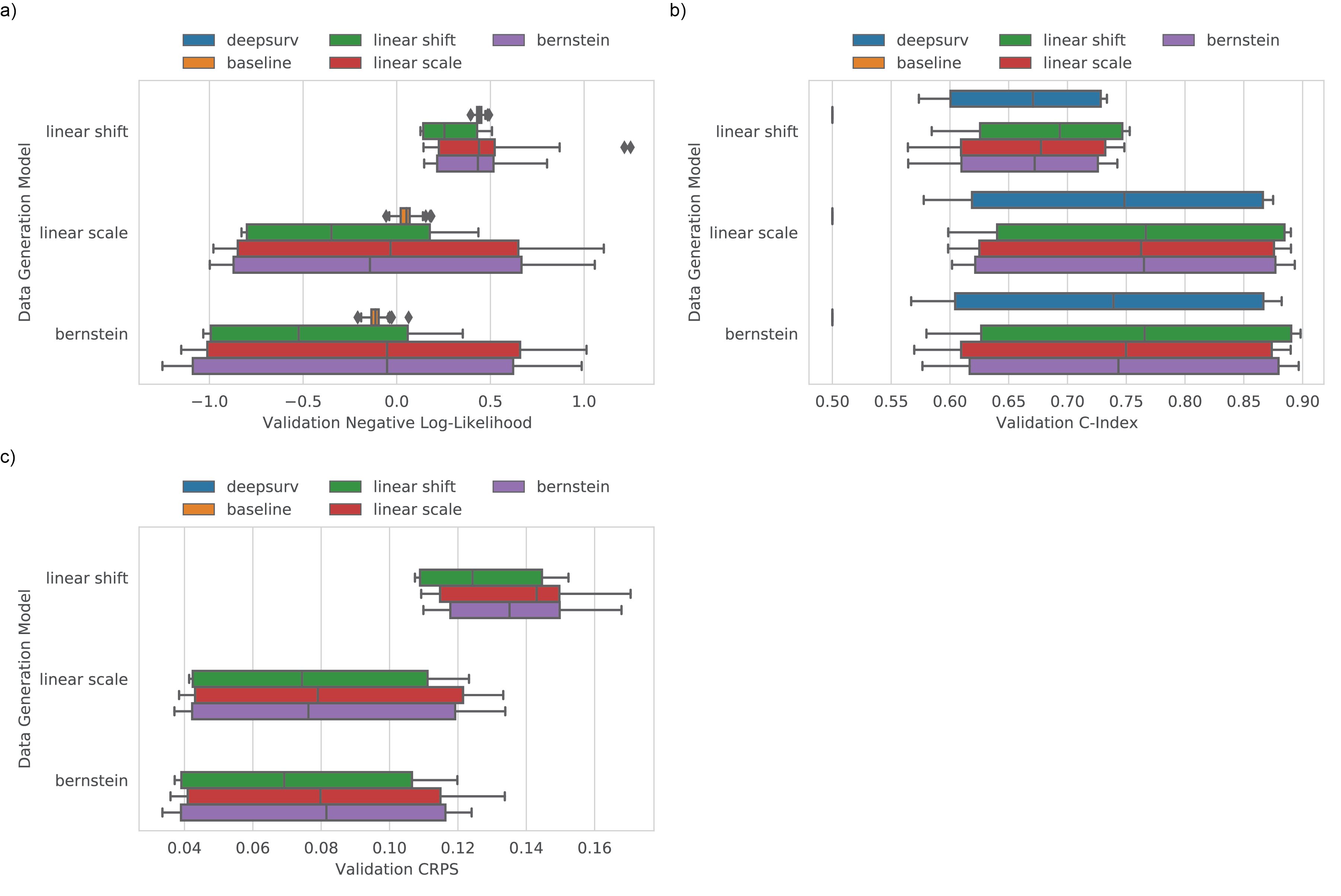}
    \caption{Results of the experiments performed on the semi-synthetic data. For each of the three 
    datasets, the performance of different parametrizations of DCTM are shown compared to DeepSurv 
    where possible. a) Distribution of validation NLL. b) Distribution of validation c-index. 
    c) Distribution of validation CRPS.}
    \label{fig:results_synth}
\end{figure}

To analyze the behavior of DCTMs when the interaction between inputs and survival comes from 
different distributions, three semi-synthetic datasets of increasing complexity were generated 
as described earlier. In Figure \ref{fig:results_synth} are shown the results of the experiments 
performed on the semi-synthetic data.
Figure \ref{fig:results_synth}a shows the distribution of validation resuts in terms of NLL for 
the three datasets fitted with various DCTM parametrizations. In all the experiments, the sigmoid 
function was used as CDF. While it is expected that data of lower complexity would be easier to fit, 
the opposite effect can be observed, where the linear shift generated data resulted in a higher NLL,
while the flexible bernstein generated data resulted in lower NLLs. Additionally, unlike our results 
on the real data, the simplest DCTM parametrization resulted in the lowest NLL in each of the three 
datasets. Figures \ref{fig:results_synth}b,c show similar trends in terms of c-index and CRPS 
respectively. In figure \ref{fig:results_synth}b it can be noted that DCTM achieves similar performance 
as DeepSurv on the three datasets.

\section{Discussion} \label{sec:outlook}

In this work we have introduced DCTMs as a novel framework for parametric and semiparametric survival 
analysis that can leverage the power of deep learning to learn patient stratification strategies directly 
from high-dimensional non-tabular data. Techniques such as this are necessary in the medical domain to 
advance the understanding of disease generation and progression.
Modelling time-to-event outcomes directly from non-tabular medical data such as radiology images, as presented 
here, has the potential to shed light in the biological processes underlying complex diseases such as cancer,
and change how patients are treated ultimately improving patient outcomes. Extensions to time-varying
covariates and competing risks exist for transformation models with tabular data \citep{fokas2017} and can be carried
over to \dlsa{s}, which we leave for future work.

\section*{Acknowledgments}

The authors are grateful for the generous computational support given by the Warren Alpert 
foundation, and the project management support of Christina M. Virgo. GC, IH, and TF were 
supported in part through the NIH/NCI Cancer Center Support Grant (grant number P30 CA008748).
The research of LK was supported by Novartis Research Foundation (FreeNovation~2019) and by 
the Swiss National Science Foundation (grant no. S-86013-01-01 and S-42344-04-01).
TH was supported by the Swiss National Science Foundation (SNF) under the project
``A Lego System for Transformation Inference'' (grant no. 200021\_184603).
TF is a founder, equity owner, and Chief Scientific Officer of Paige.AI.


\appendix
\renewcommand{\thesection}{\Alph{section}}
\counterwithin{figure}{section}
\renewcommand\thefigure{\thesection\arabic{figure}}
\counterwithin{table}{section}
\renewcommand\thetable{\thesection\arabic{table}}
\section{Notation} \label{app:notation}

$T \given X = x$ denotes a positive real-valued random variable conditional on 
covariates $X \in \mathcal{X}$. By $\phi_\theta: \mathcal{X} \to \mathbb{R}^d$, we
denote a feature extractor, \eg a convolutional neural network. We write $\phi$ and
suppress the dependency of $\phi_\theta$ on network parameters $\theta$. 
$F_{T \given X = x}$ denotes the conditional cumulative distribution function of 
$T \given X = x$, and $F_Z$ the target distribution in a transformation model.
$F_{T\given X = x}(t) = F(h(t \given x))$ then denotes a general transformation model
with transformation function $h$. For $F_Z$ we commonly chose the standard logistic 
distribution, $\sigma(z) = (1 - \exp(-z))^{-1}$, or the standard minimum extreme value 
distribution $F_{\text{MEV}}(z) = 1 - \exp(-\exp(z))$. For hazard functions we reserve 
$\lambda(t \given x)$, and $\Lambda(t \given x) = \int_0^t \lambda(u \given x)~\dd u$ 
for cumulative hazard functions.

We denote $n$ observations by $\{(t_i, x_i, \delta_i)\}_{i=1}^n$, where $t_i$ are i.i.d. 
realizations of $T \given X = x_i$, with event indicator $\delta_i \in \{0, 1\}$, where 0 
represents a right-censored and 1 an exact event time. The likelihood function is denoted by 
$\mathcal{L}(h; t, x, \delta)$, and the log-liklihood by $\ell(h; t, x, \delta)$. Depending 
on the parameterization of the transformation $h$, we substitute the parameters of $h$
as the argument of the (log-)likelihood function, \eg $\ell(a, b, w, \phi; t, x, \delta)$
for $h(t \given x) = a + bt + \phi(x)^\top w$.


\vskip 0.2in
\bibliographystyle{plainnat}
\bibliography{lk-bib} 

\end{document}